\documentclass[final,3p,times,twocolumn]{elsarticle}
\usepackage{amssymb}
\usepackage{amsmath}
\usepackage{wrapfig} 
\usepackage[ruled,linesnumbered]{algorithm2e}
\usepackage{multirow}
\usepackage[table]{xcolor} 
\usepackage{amssymb}
\usepackage{amsfonts}
\usepackage{graphicx}
\usepackage{booktabs}
\usepackage[T1]{fontenc}
\usepackage[english]{babel}
\usepackage[breaklinks=true,bookmarks=true,pagebackref=false,colorlinks,linkcolor=blue,anchorcolor=red,citecolor=blue,urlcolor=blue]{hyperref}

\usepackage{comment}
\usepackage{subcaption}
\usepackage{amsmath}
\usepackage{algorithmic}
\usepackage{array}
\usepackage[switch]{lineno}
\usepackage{longtable}
\usepackage{xspace}
\usepackage{url}
\usepackage{overpic}
\usepackage{ragged2e}
\usepackage{framed}
\usepackage{enumitem}
\usepackage{balance}
\usepackage{float}
\usepackage{makecell}
\journal{Information Fusion}

\makeatletter
\DeclareRobustCommand\onedot{\futurelet\@let@token\@onedot}
\def\@onedot{\ifx\@let@token.\else.\null\fi\xspace}

\makeatother

\definecolor{darkgreen}{rgb}{0,0.7,0}
\definecolor{darkblue}{RGB}{31,119,180}
\definecolor{darkred}{RGB}{214,39,40}
\definecolor{mediumgray}{rgb}{0.5,0.5,0.5}
\definecolor{mediumteal}{rgb}{0,0.5,0.5}

\definecolor{ellisred}{rgb}{0.87,0.44,0.38} %
\definecolor{ellisgreen}{rgb}{0.69,0.90,0.52} %
\definecolor{elliscyan}{rgb}{0.29,0.77,0.74} %
\definecolor{ellisorange}{rgb}{0.89,0.55,0.28} %
\definecolor{ellisblue}{rgb}{0.41,0.61,0.86} %

\begin{document}

\begin{frontmatter}



\title{MindDrive: An All-in-One Framework Bridging World Models and Vision-Language Model for End-to-End Autonomous Driving}

\author[1]{Bin Sun}
\author[2,3]{Yaoguang Cao}
\author[1]{Yan Wang}
\author[1]{Rui Wang}
\author[1]{Jiachen Shang}
\author[1]{Xiejie Feng}
\author[4]{Jiayi Lu}
\author[7]{Jia Shi}

% ---- Three corresponding authors ----
\author[1]{Shichun Yang\corref{cor*}}
\author[5]{Xiaoyu Yan\corref{cor*}}
\author[6]{Ziying Song\corref{cor*}}

\cortext[cor*]{Corresponding authors: yangshichun@buaa.edu.cn; yanxiaoyu@buaa.edu.cn; songziying@bjtu.edu.cn}

% ------------------ Affiliations -------------------

\address[1]{School of Transportation Science and Engineering, Beihang University}
\address[2]{State Key Laboratory of Intelligent Transportation System, Beihang University}
\address[3]{Hangzhou International Innovation Institute, Beihang University}
\address[4]{Contemporary Amperex Technology Co., Limited (CATL)}
\address[5]{Research Institute of Aero-Engine, Beihang University}
\address[6]{School of Computer Science and Technology, Beijing Jiaotong University}
\address[7]{China Automotive Engineering Research Institute Co., Ltd.}

%% Abstract
\begin{abstract}
End-to-End autonomous driving (E2E-AD) has emerged as a new paradigm, where trajectory planning plays a crucial role. Existing studies mainly follow two directions: \textbf{trajectory generation-oriented}, which focuses on producing high-quality trajectories with simple decision mechanisms, and \textbf{trajectory selection-oriented}, which perform multi-dimensional evaluation to select the best trajectory yet lack sufficient generative capability. In this work, we propose MindDrive, a harmonized framework that integrates high-quality trajectory generation with comprehensive decision reasoning. It establishes a structured reasoning paradigm of “what-if simulation – candidate generation – multi-objective trade-off.”  In particular, the proposed \textbf{Future-aware Trajectory Generator (FaTG)}, based on a \textbf{World Action Model} (WAM), performs ego-conditioned “what-if” simulations to predict potential future scenes and generate foresighted trajectory candidates. Building upon this, the \textbf{VLM-oriented Evaluator (VLoE)} leverages the reasoning capability of a large \textbf{vision–language model} to conduct multi-objective evaluations across safety, comfort, and efficiency dimensions, leading to reasoned and human-aligned decision-making. Extensive experiments on the NAVSIM-v1 and NAVSIM-v2 benchmarks demonstrate that MindDrive achieves state-of-the-art performance across multi-dimensional driving metrics, significantly enhancing safety, compliance, and generalization, and offering a promising path toward interpretable and cognitively guided autonomous driving.
\end{abstract}

%% Keywords
\begin{keyword}
Autonomous Driving\sep End-to-End Autonomous Driving\sep World Model\sep Vision Language Model
\end{keyword}

\end{frontmatter}


\section{Introduction}\label{sec:introduction}

In recent years, End-to-End autonomous driving (E2E-AD) has achieved remarkable progress and has become one of the most promising frontiers in autonomous driving research. This paradigm unifies perception, prediction, and planning within a differentiable and integrated framework, effectively reducing error propagation and improving cross-task coordination. Early research mainly focused on the perception side~\cite{chen2022learning, huang2023multi,TransFuser,shi2025fusion} to improve scene understanding, while Think Twice~\cite{thiktwice} shifted attention to the planning side, revealing that the architecture, capacity, and reasoning ability of the planning decoder are equally vital to overall system performance. As a pioneering work, UniAD~\cite{uniad} incorporates multiple sub-tasks into a single training pipeline via a dense perception–prediction–planning cascade. Subsequently, a series of lightweight E2E-AD methods employ vectorized or sparse representations to efficiently model lanes, intersections, and dynamic agents, thereby reducing computational costs. Recent studies further shifted attention toward the planning stage, coupling various generative paradigms with Transformer-based planning heads that share spatio-temporal features to improve trajectory quality. Overall, as perception modules in E2E-AD systems become increasingly mature, research focus has gradually moved from scene understanding toward efficient, safe, and comfortable trajectory planning, making planning a central topic in current E2E-AD research.

\begin{figure}
    \centering
    \includegraphics[width=\linewidth]{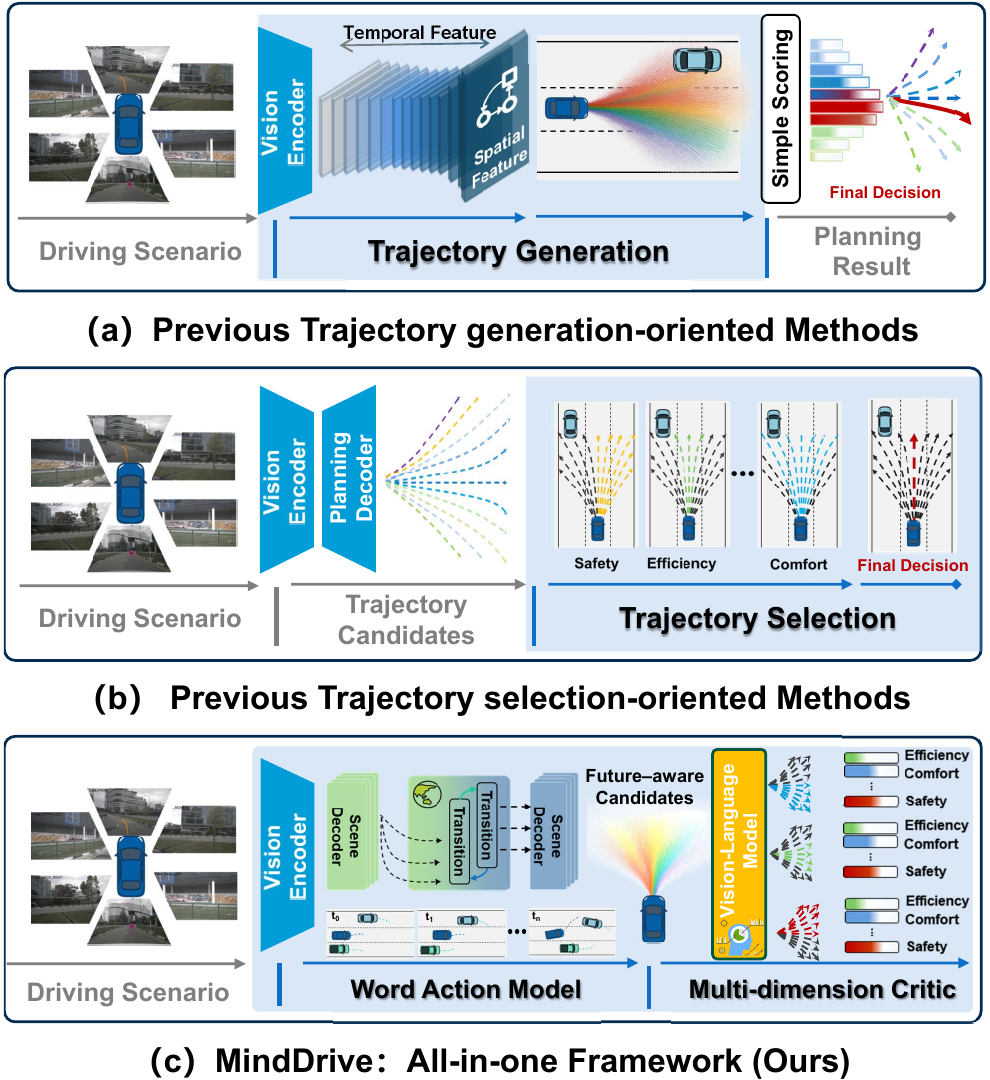}
    \caption{\textbf{Comparison between previous paradigms and our MindDrive framework.} 
    \textbf{(a)}  Trajectory generation-oriented methods~\cite{uniad,chen2024vadv2,sun2025sparsedrive,song2025don} invest heavily in producing diverse trajectories but rely on weak selectors—typically simple MLP or softmax heads—leading to suboptimal final decisions.
    \textbf{(b)}  Trajectory selection-oriented methods~\cite{li2024hydra,wote,zheng2025simplevsf} provide multi-metric evaluation (safety, efficiency, comfort) yet depend on limited candidate generation, limiting overall planning performance.
    \textbf{(c)} Our proposed MindDrive integrates world-model–based trajectory generation with VLM-driven multi-objective reasoning, achieving high-quality generation and comprehensive selection in a harmonized design.}
    \label{fig:motivation}
    \vspace{-10pt}
\end{figure}

Existing planning studies can be grouped into two major lines of work according to their focus: trajectory generation and trajectory selection. \textbf{Trajectory generation-oriented approaches}, as shown in Fig.~\ref{fig:motivation}(a), aim to improve the diversity and adaptability of planned trajectories. Representative works such as VADv2~\cite{chen2024vadv2}, GaussianAD~\cite{zheng2024gaussianad}, and DiffusionDrive~\cite{liao2025diffusiondrive} employ multimodal autoregressive modeling, probabilistic distributions, or diffusion mechanisms to capture the stochasticity of driving behaviors and complex traffic dynamics. Beyond multimodality, recent studies have further highlighted the importance of temporal coherence in trajectory generation. Methods like Bridging Past and Future~\cite{zhang2025bridging} and MomAD~\cite{song2025don} leverage historical motion cues to improve temporal stability and behavioral consistency, ensuring alignment with long-term driving intent. Moreover, the emergence of large language and vision–language models offers a new generation paradigm, enabling the incorporation of high-level semantics and scene reasoning directly into trajectory generation~\cite{Senna,xing2025openemma,zhou2025autovla}. On the other hand, \textbf{Trajectory selection-oriented approaches}, as illustrated in Fig.~\ref{fig:motivation}(b), focus on assessing pre-generated trajectory candidates to identify the most feasible one. Early works such as WoTE~\cite{wote} and Hydra-MDP~\cite{li2024hydra} train neural scoring models to mimic dataset-level benchmark metrics, using supervised learning to approximate surrogate scores defined by large driving datasets and thereby replacing hand-crafted rule-based evaluators. More recently, studies such as SimpleVSF~\cite{zheng2025simplevsf} extend this paradigm by introducing VLM-enhanced scoring, in which a vision–language model augments neural scoring networks with richer contextual and semantic understanding of the driving scene. In summary, these approaches have greatly expanded the modeling capacity of modern planning systems and laid a solid foundation for trajectory decision making. However, \textbf{Trajectory generation-oriented methods} dedicate substantial modeling capacity to crafting high-quality candidates, yet the final selection is typically performed by simple MLP heads, heuristics, or coarse scoring strategies. Such weak selectors lack multi-objective evaluation across safety, comfort, efficiency, and rule compliance. As a consequence, even when an excellent trajectory exists in the candidate set, the system may still fail to choose it, leading to sub-optimal decisions. Conversely, \textbf{Trajectory selection-oriented methods} invest heavily in sophisticated scorers but rely on traditional or lightweight mechanisms for generating the candidate trajectories themselves. In this case, regardless of how expressive the scorer is, its performance is fundamentally constrained by the coverage and quality of the candidate space—\textbf{a strong evaluator cannot rescue a weak generator}. Overall, although current methods follow the same "generate–then–select" planning pipeline, existing research typically focuses on only one side of the process, leading to \textbf{an inherent imbalance between the two stages—a core limitation of existing planning pipelines}. Consequently, even with strong generative models or powerful evaluators, E2E-AD systems often fall short of fully exploiting unified modeling and joint decision making, leaving substantial room for a more balanced and integrated planning framework.

 In complex driving scenarios, highly cognitive human drivers reason in a structured, forward-looking manner—generating candidate options, simulating their future evolutions and risks, and selecting the optimal one through multi-objective trade-offs. Achieving human-like reasoning in autonomous driving requires both the ability to anticipate how the scene may evolve, and the reasoning ability to compare multiple behavioral options in a structured manner. The introduction of world models~\cite{zheng2024occworld,zheng2025world4drive, lienhancing} offers a promising pathway for such predictive scene imagination, enabling the system to explore how future situations might unfold. Meanwhile, large language models offer complementary potential for their reasoning ability, providing high-level semantic and contextual reasoning that supports more informed evaluation of candidate behaviors~\cite{xing2025openemma,liu2024llavanext,qwen,touvron2023llamaopenefficientfoundation}.

In this work, we propose \textbf{MindDrive}, a harmonized planning framework that incorporates foresighted trajectory generation and comprehensive candidate trajectory selection. For trajectory generation, we introduce a \textbf{World Action Model (WAM)} that performs “what-if” simulations for each candidate, dynamically predicting its potential future scenarios and safety risks. This enables a \textbf{Future-aware Trajectory Generator (FaTG)} that produces more reliable and risk-sensitive candidate trajectories. For trajectory selection, we utilize the reasoning and generalization capabilities of VLM to construct a learnable multi-objective evaluation mechanism, the \textbf{VLM-oriented Evaluator (VLoE)}, which comprehensively assesses each candidate trajectory and identifies the optimal plan.

In general, our contributions are as follows.
\begin{itemize}

\item We propose the MindDrive framework, which brings together high-quality trajectory generation and comprehensive decision reasoning in a harmonized design.

\item We introduce a \textbf{Future-aware trajectory generator (FaTG)} powered by a \textbf{World Action Model (WAM)}, which simulates the spatio-temporal evolution of candidate trajectories and enables predictive, foresighted planning.

\item We design a \textbf{VLM-oriented Evaluator (VLoE)}, which leverages the semantic reasoning and generalization capabilities of vision–language models to assess candidate trajectories across multiple dimensions, including safety, comfort, and efficiency.

\item Extensive experiments on the NAVSIM-v1 and NAVSIM-v2 benchmarks demonstrate strong performance improvements over state-of-the-art E2E-AD methods.

\end{itemize}

\section{Related Work}\label{sec:related_work}
\subsection{End-to-End Autonomous Driving}
With the progressive maturity of perception technologies in End-to-End autonomous driving, the research focus has gradually shifted toward the planning side. ThinkTwice~\cite{thiktwice} emphasizes that the design of the planning module is decisive for overall system performance, which has motivated a series of systematic explorations centered on planning. Recent studies have advanced along three interconnected directions. The first line of work seeks to achieve a deeper coupling between prediction and planning~\cite{madjid2025trajectory}. For instance, SparseDrive~\cite{sun2025sparsedrive} adopts a parallel architecture that jointly optimizes prediction and planning. PPAD~\cite{chen2024ppad} also conducts iterative updates between the intentions of surrounding agents and the plan of the ego vehicle at each time step, compensating for the lack of feedback and game-theoretic interaction inherent in cascade pipelines. Building upon this foundation, a second line of research focuses on multimodal trajectory generation. VADv2~\cite{chen2024vadv2} reformulates trajectory generation from single-output regression into a discrete “vocabulary and sampling distribution” paradigm, enriching the diversity of candidate trajectories. DiffusionDrive~\cite{liao2025diffusiondrive} leverages the multimodal nature of diffusion models to enhance trajectory diversity and introduces a truncated diffusion strategy to mitigate inference inefficiency. On the other hand, GoalFlow~\cite{xing2025goalflow} adopts flow matching with goal-point mechanisms to further improve multimodal trajectory quality and consistency. The third line of work focuses on ensuring temporal coherence in planning. MomAD~\cite{song2025don} introduces trajectory and perception momentum mechanisms to produce smoother and more executable plans, while BridgeAD~\cite{zhang2025bridging} explicitly models both historical and future contexts through multi-step temporal queries, thereby strengthening the temporal consistency of planning within End-to-End frameworks.

\subsection{World Model for End-to-End Autonomous Driving}

The world models aim to predict how the environment evolves by explicitly modeling temporal dynamics, allowing the system to effectively “rehearse the future” before actual execution. Broadly, by modeling space, the existing approaches can be categorized into three streams. The first emphasizes 3D occupancy. OccWorld~\cite{zheng2024occworld} adopts 3D semantic occupancy as a unified representation, discretizes scenes into tokens via VQ-VAE, and employs an autoregressive Transformer for temporal rollout, from which the ego trajectory is decoded. A second stream simulates future states in BEV space. WoTE~\cite{wote} jointly feeds the current BEV state and multiple candidate trajectories into a world model to recursively predict future BEV semantics,  which is integrated into an online reward model to score and select trajectories. The third stream models the world in latent space. LAW (Latent World Model)~\cite{lienhancing} predicts the next-step latent representation from current scene features and ego actions, jointly optimizing latent feature learning and trajectory prediction in a self-supervised manner. World4Drive~\cite{zheng2025world4drive} builds a latent-world model based selector that forecasts future latent states for each candidate and chooses the control trajectory by comparing predicted latents with actual/reconstructed latents. Think2Drive~\cite{li2024think2drive} further adopts a latent world model within a model-based RL framework, using the world model for environment transition learning so that it can serve as both an intrinsic reward and a fast simulator for training the planner. Raw2Drive~\cite{yang2025raw2drive} builds upon Think2Drive with a dual-stream latent world model that aligns privileged features and raw-sensor dynamics. The aligned raw model functions as both an intrinsic reward source and a fast simulator, enhancing training efficiency and generalization in model-based reinforcement learning.

\subsection{LLM-based Model for End-to-End Autonomous Driving}

LLM-based models~\cite{touvron2023llamaopenefficientfoundation, liu2024llavanext, qwen}—built upon large language models (LLMs)-embed advanced reasoning capabilities, enabling more comprehensive traffic-scene understanding and improved interpretability for E2E-AD. Their strong generalization to long-tail scenarios has attracted growing interest. In E2E-AD, LLM-based models are primarily instantiated in two forms: \textbf{vision-language models (VLM)} and \textbf{vision–language–action models (VLA)}. Within VLM-augmented E2E-AD systems, three integration patterns have emerged, characterized by different levels of coupling between reasoning and planning. In hierarchical designs, for example, Senna~\cite{Senna} feeds high-level meta-actions or semantic plans produced by a VLM (e.g., “accelerate,” “turn left”) into an End-to-End planner (e.g., VADv2-style) to synthesize precise trajectories. Related works, including LMDrive and other zero-shot LLM-assisted schemes, inject periodic high-level instructions without closed-loop fine-tuning to enhance generalization. Collaborative designs, inspired by the left–right-brain synergy in human cognition, let a VLM act as a deliberative module. FASIONAD++~\cite{qian2024fasionad} triggers the VLM reasoning as a slow system to constrain the fast E2E-AD planner and mitigate information bottlenecks. In contrast, systems such as DriveVLM~\cite{qian2025agentthink} run VLM and E2E-AD branches in parallel to enhance efficiency and provide redundant safety. In holistic designs, ORION~\cite{fu2025orion} introduces a generative planner that conditions directly on VLM semantic reasoning and aligns it with the trajectory action space, enabling conditional trajectory generation. Extending VLM toward direct action generation, VLAs close the loop by mapping vision and language directly to actions, allowing an LLM-based policy to output waypoints or control inputs. OpenEMMA~\cite{xing2025openemma} serves as an early representative, bridging vision–language understanding and action control through an LLM policy that translates multimodal context into driving actions. OmniDrive~\cite{wang2025omnidrive} further advances from visual–language understanding to action production by integrating 3D scene modeling, language reasoning, and trajectory generation, leveraging expert trajectories for supervised training and QA annotation. AutoVLA~\cite{zhou2025autovla} discretizes continuous trajectories into action tokens to form clusters of candidate trajectories, then applies a reinforcement-learned policy score to select the optimal candidate for vehicle control.

\begin{figure*}[t]
\centering
 \includegraphics[width=1.0\linewidth]{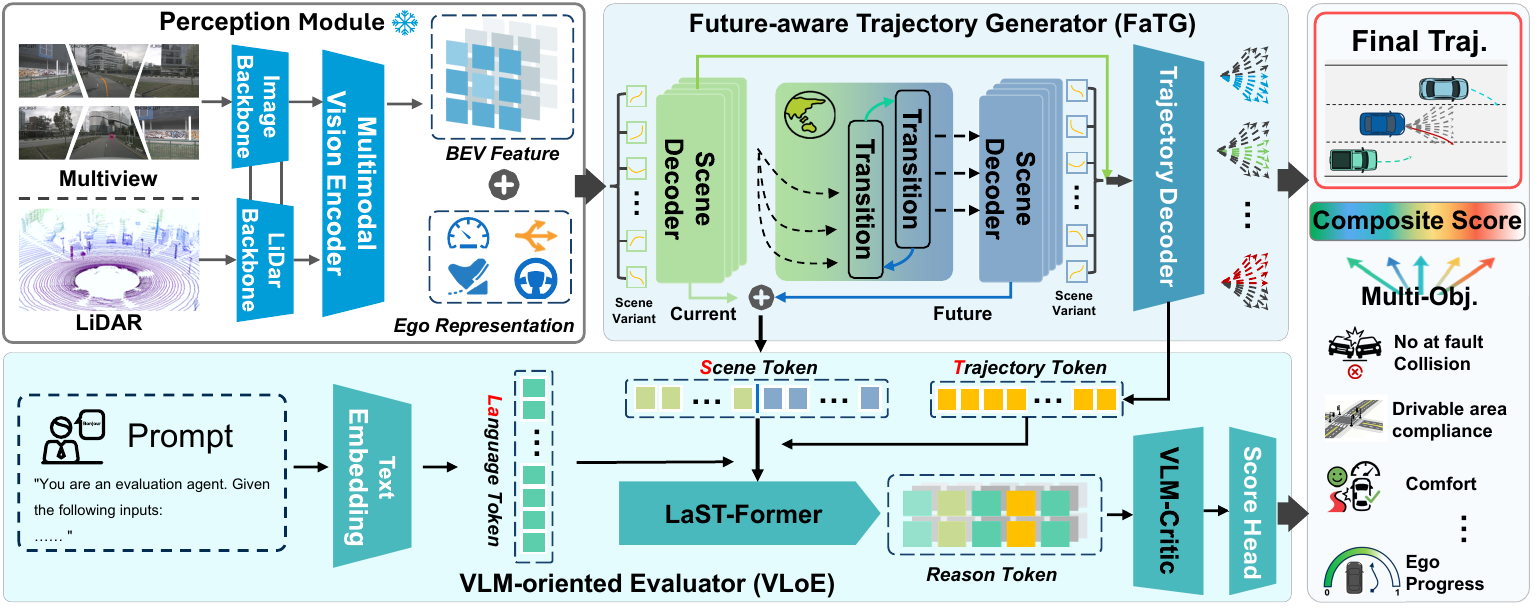}
\caption[ ]{\textbf{Overview of the MindDrive framework.} MindDrive integrates world models and vision–language models, bringing together high-quality trajectory generation and comprehensive decision reasoning in a harmonized design.\textbf{In the perception module}, multi-view camera and LiDAR inputs are fused into BEV features, while ego states and initial action intents are extracted as an \textbf{Ego Representation}. The \textbf{Future-aware Trajectory Generator (FaTG)} embeds the Ego Representation into the BEV features to construct \textbf{scene variants}, and then performs “what-if’’ simulations over them using a \textbf{World Action Model (WAM)} to model and predict their future evolutions. Subsequently, the \textbf{VLM-oriented Evaluator (VLoE)} first uses the \textbf{LaST-Former} to process multimodal tokens from the prompt and from the FaTG, generating a reasoning token. It then applies a VLM-Critic to score each trajectory on safety, comfort, efficiency, and compliance. The final trajectory is selected based on the aggregated multi-objective score.
}
\vspace{-10pt}
\label{fig:MindD}
\end{figure*}

\section{Methodology}\label{sec:method}
\subsection{Overall Architecture} 
As illustrated in Figure~\ref{fig:MindD}, we propose MindDrive, an all-in-one framework that unifies world modeling and a vision–language model for trajectory generation and evaluation in E2E-AD. The framework comprises three key stages. 

\noindent\textbf{Perception Module.} Following TransFuser~\cite{TransFuser}, we leverage multimodal sensor inputs, including multi-view RGB images and LiDAR point clouds. Visual and LiDAR features are extracted by individual backbones and fused through a multimodal encoder to form a BEV feature representation. At this stage, the ego representation is extracted, including the vehicle state—such as driving command, velocity, and acceleration—and the initial action intent derived from trajectory anchors. This representation characterizes the current driving condition and provides auxiliary cues for constructing the scene-variant representation.

\noindent\textbf{Trajectory Generation.} The \textbf{Future-aware Trajectory Generator (FaTG)} comprises a \textbf{\emph{World Action Model}} (WAM) and a \textbf{\emph{Trajectory Decoder}}. Each scene variant is constructed from the current BEV feature under different action intents. Conditioned on these variants, the WAM performs “what-if” rollouts to evolve plausible future scene states. The Trajectory Decoder subsequently integrates current and predicted features to produce diverse trajectory candidates.

\noindent\textbf{Trajectory Selection.} The core mechanism is the \textbf{Vision–Language–oriented Evaluator (VLoE)}, which critiques trajectory candidates and assigns multi-objective scores. VLoE comprises a \textit{\textbf{LaST-Former}} and a \textit{\textbf{VLM-Critic}}: the LaST-Former aligns language, scene, and trajectory tokens into \emph{reasoning tokens}, which the VLM-Critic processes via language-guided inference to produce objective-specific indicator tokens (e.g., safety, comfort, drivability). A lightweight score head aggregates these indicators into a composite score, from which the final trajectory is selected.

\subsection{Future-aware Trajectory Generator}
In most planning pipelines, generating candidate trajectories plays a pivotal role, as higher-quality candidates generally lead to more reliable decisions. However, prior methods typically rely on current-state–only predictions of surrounding agents during candidate generation, failing to account for ego-conditioned future evolution—that is, how the environment would change under alternative ego decisions.
To enhance candidate quality, we introduce a world model that allows the system to simulate plausible future scenarios, to enable the generator with explicit foresight. Accordingly, the \textbf{Future-Aware Trajectory Generator (FaTG)} comprises two key modules: a \textbf{World Action Model (WAM)}, which simulates scene evolution conditioned on different ego actions, and a \textbf{Trajectory Decoder}, which generates planning trajectories by jointly leveraging the current and predicted future states.

\subsubsection{World Action Model}

\begin{figure}[t]
\centering
 \includegraphics[width=1.0\linewidth]{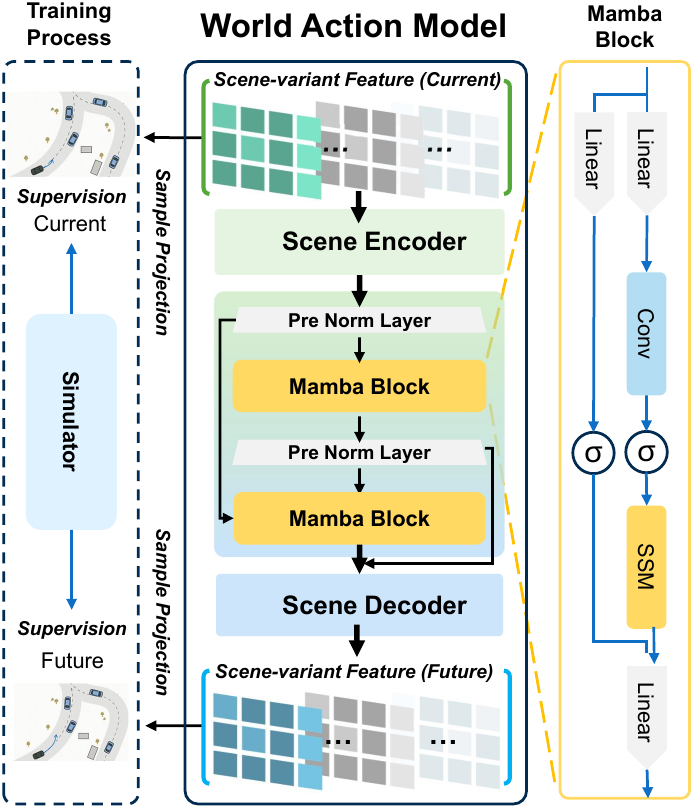}
\caption[ ]{\textbf{World Action Model (WAM).} The module follows a \textbf{spatial–temporal–spatial sandwich} design, where the Transformer models spatial dependencies and Mamba captures temporal dynamics, enabling progressive spatial encoding, temporal rollout, and spatial reconstruction of future scene representations. During training, WAM is supervised by current and future BEV semantic map features generated from the simulator, which serve as ground-truth scene-variant representations for scene rollout learning.
}
\label{fig:wam}
\end{figure}

To enable future-aware trajectory generation, we introduce a world action model, as shown in Fig.~\ref{fig:wam}, capable of simulating scene evolution and predicting diverse outcomes.

\noindent\textbf{Scene-Variant Feature Construction.} 
To represent multiple hypothetical evolutions from the same current state, we construct diverse \emph{scene-variant features} conditioned on different ego intents. 
Specifically, a set of $N$ trajectory anchors is derived by applying K-Means clustering~\cite{chen2024vadv2,li2024hydra} to all expert trajectories in the training dataset. 

We employ a self-attention encoder to process the features of each anchor. The resulting latent representation is then concatenated with the ego-status feature and passed through a MLP to produce the action token $\mathbf{a}^{(n)}_\text{token}$, which encodes the ego vehicle’s initial intent.

Let $\mathbf{B}_{\text{feature}} \in \mathbb{R}^{H \times W \times C}$ denote the BEV feature map extracted from the multimodal encoder,
where $H$ and $W$ represent the spatial height and width of the BEV grid, and $C$ is the number of feature channels.
To inject the action intent into this spatial representation, each action token $\mathbf{a}^{(n)}_\text{token}$ is distributed over $\mathbf{B}_{\text{f}}$ via bilinear interpolation.
Given the ego-centric coordinates $(x, y)$ of a point along the anchor trajectory, we project them to BEV pixel indices as:
\begin{equation}
    h = x \tfrac{H}{L_x}, \quad 
    w = y \tfrac{W}{L_y} + \tfrac{W}{2},
\end{equation}
where $L_x$ and $L_y$ denote the longitudinal and lateral BEV coverage in meters, respectively. 
The action token $\mathbf{a}_\text{token}$ is then injected into its four neighboring pixels with bilinear weights:
\begin{equation}
\mathbf{s}^{(n)}_{f}(h_i, w_j) \leftarrow 
\mathbf{B}_{\text{f}}(h_i, w_j) + 
w_{ij}\mathbf{a}^{(n)}_{\text{token}}, \quad i,j \in \{0,1\},
\end{equation}
where $w_{ij}$ denotes the bilinear interpolation weight computed from the relative distance between the ego continuous position $(h, w)$ and its four surrounding discrete grid points, and the superscript $n$ indexes the $n$-th action token being projected onto the BEV feature map.

Formally, this process can be written as:
\begin{equation}
\mathbf{s}_{f}^{(n)} = \Phi(\mathbf{B}_{f}, \mathbf{a}_\text{token}),
\end{equation}
where $\Phi(\cdot)$ denotes the bilinear injection function that fuses the ego’s action token $\mathbf{a}_\text{token}$ into the BEV feature map $\mathbf{s}_{f}$.

After injecting all action tokens, we obtain a tensor of \textbf{scene-variant features}:
\begin{equation}
\mathbf{S}_{f} = 
[\, \mathbf{s}_{f}^{(1)}, 
    \mathbf{s}_{f}^{(2)}, 
    \dots, 
    \mathbf{s}_{f}^{(N)} \,] 
\in \mathbb{R}^{N \times H \times W \times C},
\end{equation}
where each $\mathbf{s}_{f}$ corresponds to a hypothetical scene state conditioned on a specific ego intent $\mathbf{a}_\text{token}$. 
These scene-variant features serve as inputs for subsequent world-model–based scene evolution.

\noindent\textbf{World-Model Rollout.} To simulate the evolution of future driving scenarios, we perform a \textbf{World-Model rollout} based on a \textbf{spatial–temporal–spatial sandwich architecture}, which first encodes spatial dependencies, then conducts temporal reasoning over scene dynamics, and finally refines the spatial representations of future scenes.

During the rollout process, the \textbf{scene-variant features} are first processed by a Transformer encoder that captures the dynamic interactions between the ego vehicle and its surrounding environment within the scene representation.
The resulting latent representation at time step $t$ encodes a coherent spatial scene state:
\begin{equation}
\mathbf{z}_{t} = \operatorname{Transformer}_{Enc}\left(\mathbf{S}_{f,t}\right).
\end{equation}

Temporal reasoning is then performed using a stack of two \textbf{Mamba blocks}, each preceded by a pre-normalization layer and connected through residual shortcuts to ensure stable training and smooth temporal propagation.
Given the spatial latent feature $\mathbf{z}_{t}$ from the Transformer encoder, the temporal update within the Mamba stack is formulated as:

\begin{equation}
\begin{aligned}
\mathbf{z}^{(1)} &= \mathbf{z}_{t}
+\operatorname{Mamba}^{(1)}\big(\operatorname{PreNorm}^{(1)}(\mathbf{z}_{t})\big), \\
\mathbf{z}^{(2)} &= \mathbf{z}^{(1)}
+\operatorname{Mamba}^{(2)}\big(\operatorname{PreNorm}^{(2)}(\mathbf{z}^{(1)})\big), \\
\mathbf{z}_{t+k} &= \mathbf{z}^{(2)}.
\end{aligned}
\end{equation}
where the superscripts $(1)$ and $(2)$ indicate intermediate latent states produced by the first and second Mamba blocks.

Each Mamba block adopts a selective state-space mechanism that effectively models temporal dependencies with linear computational complexity.
The pre-normalization layers stabilize gradient flow and prevent over-smoothing during multi-step rollouts, while residual connections preserve spatial coherence inherited from the Transformer features.
This hierarchical design enables the model to progressively capture temporal dynamics, providing a temporally enriched latent representation for downstream scene decoding.
Furthermore, the temporal reasoning module predicts future scene states step by step over $k$ time steps in a recurrent manner, as shown below.

Finally, another Transformer layer refines the temporally evolved state and projects it back into the scene feature space, producing the \textbf{Future Scene-variant Features}:
\begin{equation}
\mathbf{S}_{f, {t+k}} = \operatorname{Transformer}_{Dec}\left(\mathbf{z}_{t+k}\right).
\end{equation}

Through multiple rollout iterations, the model produces a sequence of evolving scene representations, each reflecting a potential future evolution. This differentiable simulation allows the planner to anticipate long-term outcomes and supports future-aware trajectory generation.

\subsubsection{Trajectory Decoder}
After obtaining the temporally rolled-out scene representations, the \textbf{Trajectory Decoder} generates candidate planning trajectories by conditioning the ego action token on contextual features from both the current and predicted future scenes.

\noindent\textbf{Scene-Feature Augmentation.}
For each candidate $n$, the decoder retrieves the corresponding scene features from the current and predicted future time steps, denoted by $\boldsymbol{s}^{(n)}_{f,t}$ and $\boldsymbol{s}^{(n)}_{f,t+k}$, and concatenates them to form an augmented spatio–temporal embedding:
\begin{equation}
\boldsymbol{s}^{(n)}_{\text{aug}} \;=\;
\mathrm{Concat}\!\left(\boldsymbol{s}^{(n)}_{f,t},\, \boldsymbol{s}^{(n)}_{f,t+k}\right),
\end{equation}
which provides rich contextual cues for trajectory reasoning.

\noindent\textbf{Trajectory Candidate Generation.}
Each action token attends to the augmented scene features via a multi-head cross-attention (MHCA) layer, and the result is decoded into offsets that refine the anchor trajectory:
\begin{equation}
\boldsymbol{\tau}_{n}
\;=\;
\boldsymbol{\tau}_{n}^{\text{anchor}}
\;+\;
\Psi_{\text{offset}}\!\Big(
\mathrm{MHCA}\!\big(\boldsymbol{a}^{(n)}_\text{token},\, \boldsymbol{s}^{(n)}_{\text{aug}}\big)
\Big).
\end{equation}
Here, $\mathrm{MHCA}(\cdot)$ denotes the multi-head cross-attention operation between the action token $\boldsymbol{a}^{(n)}_\text{token}$ (query) and the augmented scene features $\boldsymbol{s}^{(n)}_{\text{aug}}$ (key/value), while $\Psi_{\text{offset}}(\cdot)$ is a trajectory decoding head that outputs offsets in $\mathbb{R}^{T \times d_{\text{pose}}}$. 

At this decoding stage, the model leverages the spatio-temporal context provided by the current scene representation and the predicted future scenes to enhance its planning reasoning capability, and generates a set of \textit{future-aware} trajectory candidates conditioned on the initial action intent.

\subsection{VLM-oriented Evaluator}

While the trajectory decoder yields diverse candidates, assessing their rationality and safety requires high-level reasoning beyond geometric feasibility. Accordingly, we introduce the \textbf{VLM-oriented Evaluator (VLoE)} (Fig.~\ref{fig:VLoE}), which leverages language–scene–trajectory alignment for interpretable, reasoning-driven assessment. VLoE comprises two components: the \textbf{LaST-Former} for multimodal fusion and a \textbf{VLM-Critic} for language-guided evaluation.

\subsubsection{LaST-Former}
\begin{figure}[t] 
\centering
\includegraphics[width=\linewidth]{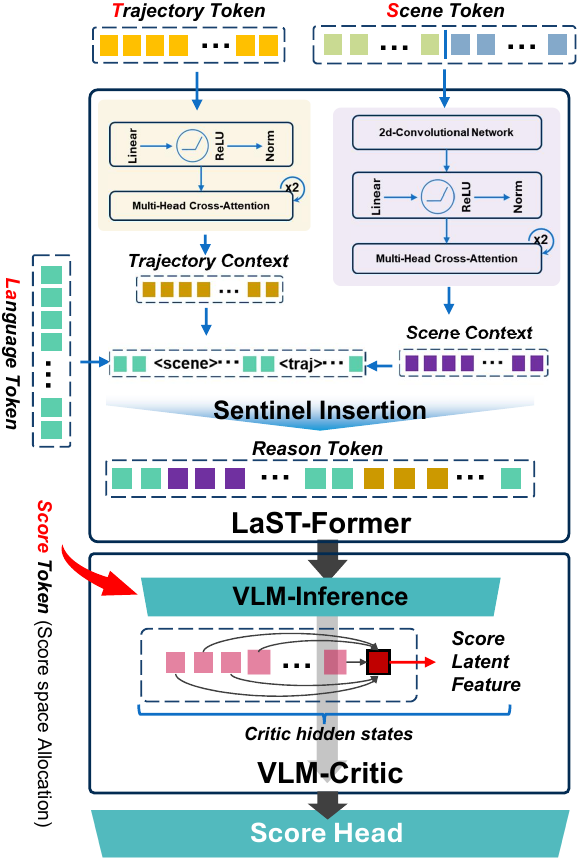}
\caption{
\textbf{VLM-oriented Evaluator (VLoE).} The \textbf{LaST-Former} fuses the language token from the prompt with
trajectory and scene tokens from the FaTG through sentinel insertion mechanism, aligning their semantics and producing a unified \textbf{reasoning tokens}. The \textbf{VLM-Critic} extends a VLM with an additional \textbf{score token} 
that aggregates scoring-related features into critic hidden states, which the score head converts into multi-objective trajectory scores.
}
\label{fig:VLoE}
\end{figure}

As illustrated in the upper part of Fig.~\ref{fig:VLoE}, the \textbf{LaST-Former} constructs \textbf{an aligned multimodal representation} by integrating scene token, trajectory token, and language cues for reasoning-driven trajectory evaluation.
It comprises two key components: \textbf{Token Encoding}, which aggregates multi-step scene features and action token into compact embeddings; \textbf{Sentinel Insertion}, which unifies this context into a reasoning sequence for downstream evaluation.

\noindent\textbf{Token Encoding.}
The scene tokens from multiple future rollout steps are concatenated along the temporal dimension to encode both spatial and temporal variations of the driving environment, forming a unified scene token:
\begin{equation}
\mathbf{S}_{\text{token}} =
\mathrm{Concat}\!\left(
\mathbf{S}_{f,t},\,
\mathbf{S}_{f,t+k}
\right).
\end{equation}
A 2D convolutional network (Conv2D) compresses the scene tokens along the spatial dimensions into a one-dimensional sequence of compact embeddings.
Subsequently, a \textbf{projection layer} composed of a lightweight \textbf{Linear–ReLU–Norm} block performs feature transformation and normalization.The resulting embeddings are further refined through a multi-head cross-attention (MHCA) block:
\begin{equation}
\mathbf{S}_\text{context} =
\operatorname{MHCA}\!\Big(
\operatorname{Proj}\!\big(
\operatorname{Conv2D}(\mathbf{S}_\text{token})
\big)
\Big)
\in \mathbb{R}^{N \times C},
\end{equation}
where $N$ denotes the number of scene tokens and $\operatorname{Proj}$ denotes an MLP head.
This process captures multi-horizon spatial dynamics and contextual consistency across rollout steps.

\noindent\textbf{Trajectory–Feature Encoding.}
In parallel, the ego representations of $K$ candidate trajectories are aggregated to form a unified trajectory token sequence:
\begin{equation}
\mathbf{T}_{\text{token}} =
\mathrm{Concat}\!\left(
\boldsymbol{\tau}_{1},
\boldsymbol{\tau}_{2},
\dots,
\boldsymbol{\tau}_{N}
\right).
\end{equation}
A \textbf{projection layer}, implemented as a lightweight \textbf{Linear–ReLU–Norm} block, maps these tokens into a compact embedding space, followed by a multi-head cross-attention (MHCA) operation for contextual refinement:
\begin{equation}
\mathbf{T}_{\text{context}} =
\mathrm{MHCA}\!\Big(
\mathbf{Q}_{\text{metric}},
\mathbf{K}=\mathbf{V}=\operatorname{Proj}(\mathbf{T}_{\text{token}})
\Big),
\end{equation}
where $\mathbf{Q}_{\text{metric}} \in \mathbb{R}^{K \times C}$ denotes a set of learnable \textit{metric queries} that attend to all candidate trajectories via cross-attention, producing compact trajectory representations.
These metric-oriented queries serve as evaluation probes, capturing motion intent and inter-trajectory dependencies that provide discriminative cues for downstream reasoning and scoring modules.

\noindent\textbf{Sentinel Insertion.}
To enable unified multimodal reasoning within the language model, we adopt a token-level integration strategy.
The fused scene and trajectory embeddings are serialized and injected into the textual sequence using predefined \textbf{sentinel tokens} (e.g., \texttt{<scene>} and \texttt{<traj>}), which indicate the positions for multimodal feature substitution.
During tokenization, textual tokens are embedded conventionally, while each sentinel is assigned a reserved index (e.g., \(-200\)) that serves as a placeholder.
During the embedding stage, these placeholders are replaced by the corresponding scene or trajectory embeddings, ensuring their alignment with the textual stream.
This mechanism allows multimodal features and linguistic representations to coexist within a unified token space, enabling cross-modal reasoning in subsequent VLM processing.

Overall, the LaST-Former aggregates multi-horizon spatial features, ego trajectory semantics, and language cues into a shared latent space, providing a scene-grounded and cognitively aligned representation for reasoning-based trajectory evaluation.

\subsubsection{VLM-Critic}
To enable language-guided trajectory evaluation, we extend the base language model’s vocabulary with a special \textbf{score token} (\texttt{<score\_feature>}). 
Unlike ordinary tokens, it aggregates contextual semantics from both textual and multimodal embeddings, serving as a dedicated reasoning node for evaluation. 
During inference, the model processes the multimodal prompt in a causal manner, and the hidden state of the score token is extracted as the evaluation feature:
\begin{equation}
\mathbf{h}_{\text{eval}} =
\mathrm{VLM}_\text{infer}\!\left(
\mathbf{T}_\text{reason},
\mathbf{m}_\text{attn}
\right)
\Big|_{\texttt{score}},
\end{equation}
where $\mathrm{VLM}_{infer}(\cdot)$ performs multimodal reasoning over the token sequence; 
$\mathbf{T}_\text{reason}$ includes the textual prompt, and multimodal sentinels aligned with scene-trajectory context; and 
$\mathbf{m}_{\text{attn}}$ denotes the attention mask controlling token visibility. 
The operator $\big|_{\texttt{score}}$ uses a designated score token to aggregate the hidden states of the entire output sequence, producing the evaluation embedding $\mathbf{h}_{\text{eval}}$.
Unlike standard text generation, this token focuses on trajectory assessment, and its hidden state is passed through a scoring head $\Psi_{\text{eval}}$ to produce quantitative evaluation results:
\begin{equation}
\mathbf{r}_{\text{score}} =
\Psi_{\text{eval}}\!\left(\mathbf{h}_{\text{eval}}\right),
\end{equation}
yielding trajectory scores based on predefined reasoning rules and assessment metrics.

\section{Experiments}\label{sec:exps}

% Table 1

\begin{table*}[t!]
    % \vspace{-0.3cm}
    \centering
    \resizebox{\textwidth}{!}{
        \begin{tabular}{l|c|c|ccccc
        | >{\columncolor{gray!25}}c}
            \toprule
            Method & Input  & Venue & NC $\uparrow$ & DAC $\uparrow$ & EP $\uparrow$ & TTC $\uparrow$ & Comf. $\uparrow$ & PDMS $\uparrow$ \\
            \midrule
            Human & - &  -   & 100 & 100 & 87.5 & 100 & 99.9 & 94.8 \\
            \midrule

            LTF~\cite{TransFuser} & C  & TPAMI 2022  & 97.4 & 92.8 & 79.0 & 92.4 & 100.0 & 83.8 \\

            TransFuser~\cite{TransFuser} & C \& L   & TPAMI 2022  & 97.7 & 92.8 & 79.2 & 92.8 & \textbf{100} & 84.0 \\

            UniAD~\cite{uniad} & C                  & CVPR 2023  & 97.8 & 91.9 & 78.8 & 92.9 & \textbf{100} & 83.4 \\
            
            PARA-Drive~\cite{weng2024drive} & C     & CVPR 2024  & 97.9 & 92.4 & 79.3 & 93.0 & 99.8 & 84.0 \\
            
            VADv2~\cite{chen2024vadv2} & C          & arXiv 2024  & 97.9 & 91.7 & 77.6 & 92.9 & \textbf{100} & 83.0 \\
            
            DRAMA~\cite{yuan2024drama} & C \& L     & arXiv 2024  & 98.0 & 93.1 & 80.1 & 94.8 & \textbf{100} & 85.5\\
             
            Hydra-MDP~\cite{li2024hydra} & C \& L   & arXiv 2024  & 98.3 & 96.0 & 78.7 & 94.6 & \textbf{100} & 86.5 \\

            Hydra-MDP++~\cite{li2025hydra} & C \& L   & arXiv 2024  & 97.6 & 96.0 & 80.4 & 93.1 & \textbf{100} & 86.6 \\
            
            LAW~\cite{lienhancing} & C              & ICLR 2025  & 96.4 & 95.4 & 81.7 & 88.7 & 99.9 & 84.6 \\
            
            GoalFlow~\cite{xing2025goalflow} & C \& L  & CVPR 2025 & 98.3 & 93.8  & 79.8 & 94.3 & \textbf{100} & 85.7 \\
            
            ARTEMIS~\cite{feng2025artemis} & C \& L  & arXiv 2025  & 98.3 & 95.1 & 81.4 & 94.3 & \textbf{100}  & 87.0 \\
            
            DiffusionDrive~\cite{liao2025diffusiondrive} 
                                           & C \& L  & CVPR 2025  & 98.2 & 96.2 & 82.2 & 94.7 & \textbf{100}  & 88.1 \\
            
            WoTE~\cite{wote} & C \& L          & ICCV 2025 & \textbf{98.5} & 96.8 & 81.9 & 94.9 & 99.9 & 88.3 \\
            
            DIVER~\cite{diver} & C \& L  & arXiv 2025 & \textbf{98.5} & 96.5 & 82.6 & 94.9 & \textbf{100} & 88.3 \\
            \midrule
            \textbf{MindDrive (Ours)} & C \& L  & -  & 98.4 & \textbf{97.0} & \textbf{82.8} & \textbf{95.1} & 99.9 & \textbf{88.9}
            \\
            \bottomrule
        \end{tabular}
    }
\caption{\textbf{NAVSIM-v1 evaluation results on the Navtest split.} 
C: Camera. L: LiDAR. NC: no at-fault collision. DAC: drivable area compliance. 
EP: ego progress. TTC: time-to-collision. Comf.: comfort. 
PDMS: the predictive driver model score. 
For fair comparison, all methods are evaluated using a unified \textbf{ResNet-34} backbone.}

\label{tab:navsimv1_sota}
\end{table*}

% Table 2
\begin{table*}[t]
    \centering
    \setlength{\tabcolsep}{3pt}
    \renewcommand{\arraystretch}{1.2}
    \resizebox{\textwidth}{!}{ 
      \begin{tabular}{l|ccccccccc
        ||>{\columncolor{gray!25}}c
        |>{\columncolor{gray!25}}c}
    \toprule
    Method 
    &NC$\uparrow$ & DAC$\uparrow$ & EP$\uparrow$ & TTC$\uparrow$ & HC$\uparrow$ &
    TLC$\uparrow$ & DDC$\uparrow$ & LK$\uparrow$ & EC$\uparrow$ & EPDMS$_1$ $\uparrow$ & PDMS $\uparrow$\\
    \midrule
    Transfuser~\cite{TransFuser} 
    
    & 97.7 & 92.8 & 79.2 & 92.8 & 100 & 99.9 & 98.3 & 67.6 & 95.3 & 77.8  & 84.0  \\
    
    VADv2~\cite{chen2024vadv2} 
    
    & 97.3 & 91.7 & 77.6 & 92.7 & 100 & 99.9 & 98.2 & 66.0 & 97.4 & 76.6  & 83.0 \\
    
    Hydra-MDP~\cite{li2024hydra}
 
    & 97.5 & 96.3 & 80.1 & 93.0 & 100 & 99.9 & 98.3 & 65.5 & 97.4 & 79.8  & 86.5 \\
    
    Hydra-MDP++~\cite{li2025hydra}
 
    & 97.9 & \textbf{96.5} & 79.2 & 93.4 & \textbf{100} & 100 & 98.9 & 67.2 & 97.7 & 80.6 & 86.6\\

    ARTEMIS~\cite{feng2025artemis} 

    & 98.3 & 95.1 & 81.5 & 97.4 & 100 & 99.8 & 98.6 & 96.5 & \textbf{98.3} & 83.1 & 87.0 \\
    \midrule
    \textbf{MindDrive (Ours)} 

    & \textbf{98.4} & 95.6 & \textbf{86.7} &\textbf{97.5} & \textbf{100} & 99.8 & \textbf{99.3} &   \textbf{96.5} & 96.8 & \textbf{84.2}  & \textbf{88.9} \\
    \bottomrule
    \end{tabular}
    }
    \caption{\textbf{NAVSIM-v2 evaluation results on the Navtest split.} NC: No at-fault Collision; DAC: Drivable Area Compliance; DDC: Driving Direction Compliance; TLC: Traffic Light Compliance; EP: Ego Progress; TTC: Time to Collision; LK: Lane Keeping; HC: History Comfort; EC: Extended Comfort; EPDMS$_1$: the Extended predictive driver model score for stage~1. PDMS: identical to the values in Table~\ref{tab:navsimv1_sota} for the same method.For fair comparison, all methods are evaluated using a unified \textbf{ResNet-34} backbone.}
    \label{tab:navsimv2_navtest_sota}
\end{table*}

% Table 3
\begin{table*}[!ht]
\centering
\setlength{\tabcolsep}{3.5pt} % 调整列间距
\renewcommand{\arraystretch}{1.1} % 调整行高
\resizebox{\textwidth}{!}{ % 表格宽度缩放
\begin{tabular}{l|l|ccccccccc
||>{\columncolor{gray!25}}c}
\toprule
Method
& Stage & NC$\uparrow$ & DAC$\uparrow$ & DDC$\uparrow$ & TLC$\uparrow$ & EP$\uparrow$ & TTC$\uparrow$ & LK$\uparrow$ & HC$\uparrow$ & EC$\uparrow$ & EPDMS$\uparrow$ \\
\midrule
% 修改点1：将引用从multirow中取出，放在方法名后
\multirow{2}{*}{LTF~\cite{TransFuser}}

& Stage I
& 96.2 & 79.5 & \textbf{99.1} & \textbf{99.5} & 83.1 & 95.1 & 94.2 & 97.5 & \textbf{79.1} &\cellcolor{gray!25} \\
& Stage II
& 77.7 & 70.2 & 84.2 & 98.0 & 85.1 & 75.6 & 45.4 & 95.7 & \textbf{75.9} & \multirow{-2}{*}{23.1} \\

\midrule
\multirow{2}{*}{DiffusionDrive~\cite{liao2025diffusiondrive}} 

& Stage I 
& 96.0 & 79.7 & 97.4 & \textbf{99.5} & 81.3 & 93.1 & 90.8 & 96.8 & 73.8 & \cellcolor{gray!25} \\
& Stage II 
& 82.1 & 72.2 & 88.5 & 98.7 & 85.1 & 78.8 & 49.2 & 89.3 & 71.2 & \cellcolor{gray!25}\multirow{-2}{*}{24.2} \\

\midrule
\multirow{2}{*}{GTRS-DP~\cite{li2025generalized}} 

& Stage I  
& 94.7 & 78.8 & 96.1 & \textbf{99.5} & 83.0 & 94.4 & 92.0 & 97.5 & 72.8 &  \\

& Stage II 
& 80.3 & 74.4 & 84.9 & 98.0 & 81.9 & 78.8 & 45.4 & 96.7 & 70.1 & \multirow{-2}{*}{23.8}  \\

\midrule
\multirow{2}{*}{GuideFlow~\cite{liu2025guideflow}} 
& Stage I  
& \textbf{96.6} & 80.5 & 96.3 & 99.3 & 82.3 & 94.9 & 91.5 & \textbf{97.7} & 67.8 &  \\
& Stage II 
& \textbf{87.3} & 76.7 & \textbf{88.8} & \textbf{99.2} & 84.3 & \textbf{85.1} & \textbf{49.7} & 93.1 & 44.5 & \multirow{-2}{*}{27.1} \\

\midrule
\multirow{2}{*}{\textbf{MindDrive (Ours)}}

  & Stage I  & 96.1 & \textbf{86.0} & 98.8 & 99.3 & \textbf{83.3} & \textbf{95.6} & \textbf{94.4} & 97.6 & 74.7 &  \\

  & Stage II & 82.6 & \textbf{79.1} & 86.4 & 98.0 & \textbf{85.3} & 79.4 & 49.2 & 96.5 & \textbf{71.0} & \multirow{-2}{*}{\textbf{30.5}}\\
\bottomrule
\end{tabular}
}
\caption{\textbf{NAVSIM-v2 evaluation results on Navhard split.} NC: No at-fault Collision; DAC: Drivable Area Compliance; DDC: Driving Direction Compliance; TLC: Traffic Light Compliance; EP: Ego Progress; TTC: Time to Collision; LK: Lane Keeping; HC: History Comfort; EC: Extended Comfort; EPDMS: the final Extended Predictive Driver Model Score obtained by combining Stage~I and Stage~II evaluations. Stage I evaluates open-loop trajectories under initial observations; Stage II repeats evaluation on synthetic future observations. The final EPDMS is obtained by Gaussian-weighted aggregation of both stages. For fair comparison, all methods are evaluated using a unified \textbf{ResNet-34} backbone.}
\label{tab:navsimv2_sota}
\end{table*}

\subsection{Dataset and metrics}
Our experiments are conducted on the \textbf{NAVSIM} framework~\cite{navsim, navsimv2}, a planning-oriented evaluation benchmark built upon OpenScene~\cite{openscene2023}, a compact redistribution of the nuPlan dataset~\cite{caesar2021nuplan}. \textbf{NAVSIM} is grounded in real-world multi-sensor driving data, inheriting the full sensor configuration of its parent datasets and providing 2 Hz annotations of multi-agent trajectories and HD maps. \textbf{NAVSIM-v1}~\cite{navsim} evaluates planners under a non-reactive open-loop protocol using the Predictive Driver Model Score (PDMS), whereas \textbf{NAVSIM-v2}~\cite{navsimv2} extends this setup to a pseudo–closed-loop scoring process (EPDMS) with synthetic future observations to better approximate closed-loop behavior. \textbf{NAVSIM} is designed to highlight challenging and safety-critical driving situations, providing a standardized and behavior-grounded benchmark for evaluating E2E-AD systems. The \textbf{Navtest split} is the standard testing dataset in NAVSIM. 

\textbf{NAVSIM-v1.} Unlike standard open-loop datasets that provide only trajectory pairs for ADE/FDE computation, NAVSIM-v1 embeds each logged scene into a lightweight, non-reactive simulation environment. Performance is measured using the \textbf{Predictive Driver Model Score (PDMS)}, a composite metric that integrates: \textbf{Safety}: No-at-fault Collision (NC), Time-to-Collision (TTC), \textbf{Compliance}: Drivable Area Compliance (DAC), \textbf{Efficiency}: Ego Progress (EP), \textbf{Comfort}: comfort-related penalty. PDMS provides a more behavior-grounded evaluation of planning quality. 

\noindent The \textbf{Navsafe split}. The Navsafe split is a safety-critical subset of the Navtest split~\cite{sima2025centaur}, constructed by combining NHTSA pre-crash typologies~\cite{najm2007pre}, human inspection, and CLIP-based clustering~\cite{ilharco2021openclip} to mine rare and challenging scenarios from the long-tail distribution. It covers ten high-risk categories—including roundabouts, unprotected left turns, ramps, yellow-light dilemmas, no-lane areas, unusual signs, bad weather, overtaking, and yielding—capturing situations that frequently cause failures in planning. 

\textbf{NAVSIM-v2.} The v2 release upgrades the scoring metric from PDMS to the \textbf{Extended Predictive Driver Model Score (EPDMS)}, with additional metrics including Driving Direction Compliance (DDC), Traffic Light Compliance (TLC), Lane Keeping (LK), and Extended Comfort (EC). 

\noindent The  \textbf{Navhard split}. Furthermore, NAVSIM-v2 introduces an additional Navhard split, which adopts a two-stage evaluation workflow. In \textbf{Stage~I}, the planner predicts a trajectory from the real observation and receives an initial score $EPDMS_1$. In \textbf{Stage~II}, multiple plausible future observations are synthesized around the Stage~1 endpoint using 3D Gaussian Splatting, and the planner is re-evaluated under these perturbed conditions to obtain $EPDMS_2$. The two scores are fused through Gaussian-weighted aggregation to produce the final score $EPDMS$.

\subsection{Implementation Details}
\label{sec:Implementation_Details}

We adopt a vision–LiDAR fusion strategy to jointly leverage semantic and geometric cues from multimodal inputs. Synchronized front, front-left, and front-right RGB images are center-cropped, aligned, and concatenated into a $256 \times 1024$ composite, while LiDAR point clouds within a $64,\text{m} \times 64,\text{m}$ range around the ego vehicle are projected onto a BEV grid. A ResNet-34~\cite{resnet} backbone encodes both modalities, whose features are extracted independently and fused via transformer-based cross-attention in BEV space, following TransFuser~\cite{TransFuser}. For high-level reasoning, we incorporate a Tiny-LLaVA-1B~\cite{zhou2024tinyllava} vision–language model as the inference backbone and fine-tune it using the LoRA strategy to efficiently adapt to driving-scene reasoning and evaluation tasks.

\subsection{Main Results}

We evaluate \textbf{MindDrive} under the NAVSIM benchmark across its three complementary evaluation protocols.
\textbf{NAVSIM-v1} provides the standard PDMS-based assessment on the Navtest split, serving as the baseline protocol.
\textbf{NAVSIM-v2 (Navtest)} uses the same scenarios but replaces PDMS with a richer and stricter set of compliance metrics, enabling a finer-grained examination of planner behavior. Finally, \textbf{NAVSIM-v2 (Navhard)} adopts a pseudo-simulation setting with synthetic futures, introducing a two-stage causal evaluation that stresses robustness under interactive and safety-critical conditions. Together, these three protocols offer a progressive view of planning performance—from overall driving quality (v1-Navtest), to fine-grained compliance (v2–Navtest), to full causal and interactive robustness (v2–Navhard).

\subsubsection{NAVSIM-v1 Evaluation Results on the Navtest split}

The Navtest split serves as the standard benchmark test set in NAVSIM, containing a fixed set of challenging real-world scenarios used for evaluating planning performance. As shown in Table~\ref{tab:navsimv1_sota}, \textbf{MindDrive} achieves state-of-the-art performance under the NAVSIM-v1 protocol. Across all PDMS components—including NC, DAC, TTC, Comfort, and EP—MindDrive consistently surpasses previous methods. In particular, it obtains a PDMS score of \textbf{88.9}, outperforming recent strong baselines such as DiffusionDrive~\cite{liao2025diffusiondrive}, WoTE~\cite{wote}, and DIVER~\cite{diver}. MindDrive further exhibits strong safety (TTC 95.1) and smooth (Comf. 99.9) and efficiency (EP 82.8) driving behavior, confirming the effectiveness of the proposed architecture under the original NAVSIM-v1 assessment.

\subsubsection{NAVSIM-v2 Evaluation Results on the Navtest split.}
To complement the NAVSIM-v1 evaluation, we additionally assess MindDrive under the stricter NAVSIM-v2 protocol on the same Navtest split. Since the split contains only real scenarios and does not support Stage~II, we report the Stage-1 score (EPDMS$_1$). Since Navtest includes only real scenarios—without the synthetic augmentations required for Stage-2—we report only the Stage~I score EPDMS$_1$ under NAVSIM-v2. Table~\ref{tab:navsimv2_navtest_sota} summarizes the results. Under the expanded and more stringent metric set of NAVSIM-v2, all baselines show noticeable declines in performance, reflecting the increased difficulty of the protocol. In contrast, \textbf{MindDrive} maintains strong results across all newly introduced metrics, achieving \textbf{99.3 DDC}, \textbf{96.5 LK}, and \textbf{96.8 EC}. Consequently, MindDrive attains an EPDMS$_1$ of \textbf{84.2}, demonstrating robust safety, feasibility, and compliance under the enhanced NAVSIM-v2 criteria.

\subsubsection{NAVSIM-v2 Evaluation results on the Navhard split} 
Table~\ref{tab:navsimv2_sota} reports results on the Navhard split, the pseudo-simulation benchmark introduced in NAVSIM-v2. \textbf{MindDrive} achieves the highest overall score with an EPDMS of \textbf{30.9}. In Stage~I, MindDrive leads across most safety-critical metrics, including \textbf{NC (96.1)}, \textbf{DAC (86.0)}, and \textbf{TTC (99.3)}, while also maintaining strong controllability and comfort (\textbf{HC 94.4}). In Stage~II—where the evaluation is repeated with synthetic future observations—MindDrive remains competitive, particularly on compliance metrics such as \textbf{DAC (79.1)} and \textbf{TLC (98.0)}. Overall, these results show that MindDrive delivers reliable and consistent driving behavior under the more challenging and interactive conditions of the Navhard benchmark.

\subsection{Robustness Study}

% Robustness Analysis
\begin{table}
    \centering
    
\centering
\addtolength{\tabcolsep}{0.2pt}
\renewcommand{\arraystretch}{0.9} %
\resizebox{\linewidth}{!}{
\begin{tabular}{l|>{\columncolor{gray!30}}c|cccccc}
\toprule
Method & 
PDMS$\uparrow$ &
NC$\uparrow$ & 
DAC$\uparrow$ & 
EP$\uparrow$ & 
TTC$\uparrow$ & 
Comf.$\uparrow$   \\
\midrule
TransFuser$^*$~\cite{TransFuser} 
& 55.8 & 95.4 & 60.9 & 51.8 & 93.2 & 98.9   \\

DiffusionDrive$^*$~\cite{liao2025diffusiondrive}
& 66.1 & 96.6 & 71.2 & 61.9 & 91.4 & 100 \\

WoTE$^*$~\cite{wote} 
& 65.3 & \textbf{98.7} & 71.9 & 58.5 & \textbf{96.3} & 99.0 \\
\midrule
\textbf{MindDrive}
& \textbf{67.5} & 98.5 & \textbf{75.6} & \textbf{63.1} & 95.5 & \textbf{100}  \\

\bottomrule
\end{tabular}}
\caption{\textbf{Robustness analysis on the Navsafe split under the NAVSIM-v1 metrics.} *: results reproduced using official weights.}
\label{tab:rob_navsafe}

    \vspace{-10pt}
\end{table}

\begin{table}
    \centering
    
\centering
\addtolength{\tabcolsep}{0.2pt}
\tabcolsep=0.5mm
\renewcommand{\arraystretch}{0.9} %
\resizebox{\linewidth}{!}{
\begin{tabular}{l|>{\columncolor{gray!30}}c|ccccccccc}
\toprule
Method & EPDMS$_1$ $\uparrow$ & NC$\uparrow$ & DAC$\uparrow$ & DDC$\uparrow$ & TLC$\uparrow$ & EP$\uparrow$ & TTC$\uparrow$ & LK$\uparrow$ & HC$\uparrow$ & EC$\uparrow$ \\
\midrule
LTF$^{*}$~\cite{TransFuser} 
& 62.3 & 96.2 & 79.5 & 99.1 & 99.5 & 84.1 & 95.1 & 94.2 & 97.5 & \textbf{79.1}   \\

Transfuser$^{\dagger}$~\cite{TransFuser} 
& 60.5 & 96.3 & 74.6 & 98.4 & 99.3 & 82.9 & 93.7 & 92.7 & 97.5 & 78.2   \\

DiffusionDrive$^{\dagger}$~\cite{liao2025diffusiondrive}
& 63.2 & 95.9 & 84.0 & \textbf{98.6} & \textbf{99.8} & \textbf{84.4} & 96.0 & 95.1 & \textbf{97.6} & 71.1 \\

WoTE$^{\dagger}$~\cite{wote} 
& 66.7 & \textbf{97.4} & 88.2 & 97.8 & 99.3 & 82.7 & \textbf{96.4} & 90.9 & 97.3 & 68.0 \\

\midrule
\textbf{MindDrive$^{\dagger}$}
& \textbf{72.7} 
& 97.0 & \textbf{89.7} & \textbf{98.6} & 99.3 & 83.7 & 95.8 & \textbf{96.0} & 97.5 & 71.1 \\

\bottomrule
\end{tabular}}
\caption{\textbf{Robustness analysis on the Navhard split under the NAVSIM-v2 metrics.} *: denotes results reproduced from the official weights.$^\dagger$ denotes methods operating on multimodal sensor inputs (camera + LiDAR).  EPDMS$_1$ represents the EPDMS stage 1 evaluated without synthetic scenarios.}
\label{tab:rob_navhard_stage_one}

\end{table}

To comprehensively assess the robustness of our framework, we further evaluate on the \textbf{Navsafe} and \textbf{Navhard} splits, using the NAVSIM-v1 and NAVSIM-v2 metrics respectively. The Navsafe split is a safety-critical subset of Navtest, containing rare high-risk scenarios mined through NHTSA pre-crash typologies and additional filtering. Additionally, Navhard is a difficult subset featured in the stricter NAVSIM-v2 leaderboard, covering unprotected turns, dense traffic, and other challenging situations.

\noindent\textbf{The Navsafe split under NAVSIM-v1 metrics.} Table~\ref{tab:rob_navsafe} shows that MindDrive achieves the highest overall robustness on the Navsafe split, obtaining a PDMS of 67.5 and consistently outperforming all reproduced baselines across safety (NC, DAC), efficiency (EP), and comfort metrics (TTC, Comf.). Since Navsafe focuses on rare high-risk but non-adversarial scenarios, these results demonstrate that our generation–selection framework maintains strong safety guarantees and planning stability under safety-critical conditions.

\noindent\textbf{The Navhard split under NAVSIM-v2 metrics (Stage~I).} Because Stage-2 evaluation in NAVSIM-v2 relies solely on \emph{camera-only} inputs, we report only the Stage-1 results to ensure a fair comparison among multimodal methods that jointly use camera and LiDAR signals. As shown in Table~\ref{tab:rob_navhard_stage_one}, \textbf{MindDrive} achieves the highest Stage-1 EPDMS score (72.7) and maintains strong performance across key safety and compliance metrics—such as \textbf{NC (97.0)}, \textbf{DAC (89.7)}, \textbf{DDC (98.6)}, and \textbf{TTC (95.8)}—despite the increased difficulty of the Navhard split. It is worth noting that, since MindDrive operates in a multimodal setting here, its Stage-I results naturally differ from those reported in Table~\ref{tab:navsimv2_sota}, which evaluates on a different sensor configuration. Overall, the ability to retain reliable driving signals under severe disturbances highlights the robustness of MindDrive’s multimodal reasoning in challenging scenarios.

\begin{table}[!t]
    \centering
        
\centering
\tabcolsep=1.2mm
\renewcommand{\arraystretch}{0.9}
\resizebox{\linewidth}{!}{
\begin{tabular}{
>{\centering\arraybackslash}p{15mm} |
>{\centering\arraybackslash}p{15mm} |
>{\columncolor{gray!30}}c|ccccc}
\toprule
FaTG & VLoE & PDMS $\uparrow$ & NC $\uparrow$ & DAC $\uparrow$ & EP $\uparrow$ & TTC $\uparrow$ & Comf. $\uparrow$ \\
\midrule
$\times$ & $\times$ & 84.1 & 97.8 & 92.5 & 79.3 & 93.0 & 99.8 \\
$\checkmark$ & $\times$ & 86.6 & 97.9 & 95.0 & 81.0 & 94.0 & \textbf{99.9} \\
$\times$ & $\checkmark$ & 87.7 & 98.1 & 96.4 & 81.7 & 94.2 & \textbf{99.9} \\
\midrule
$\checkmark$ & $\checkmark$ & \textbf{88.9} & \textbf{98.4} & \textbf{97.0} & \textbf{82.8} & \textbf{95.1} & \textbf{99.9} \\
\bottomrule
\end{tabular}
}
\caption{\textbf{Ablation study of the overall architecture.} (On the Navtest split under the NAVSIM-v1 metrics.)
FaTG: Future-aware Trajectory Generator. VLoE: VLM-oriented Evaluator.}
\label{tab:ablation_overall_PDMS}

    \vspace{-10pt}
\end{table}

\subsection{Ablation Study}

\subsubsection{Ablation Study on overall Architecture}
To investigate the contribution of each component in \textbf{MindDrive}, we conduct an ablation study on the Navtest split under the NAVSIM-v1 metrics, examining the roles of the \textbf{Future-aware Trajectory Generator (FaTG)} and the \textbf{VLM-oriented Evaluator (VLoE)}. As shown in Table~\ref{tab:ablation_overall_PDMS}, removing either module leads to clear performance drops across multiple closed-loop metrics.  Introducing \textbf{FaTG} yields consistent gains in safety and feasibility. This demonstrates the importance of future-aware trajectory generation for reliable planning. Likewise, enabling \textbf{VLoE} noticeably improves overall PDMS and boosts high-impact driving metrics such as EP and TTC. These results indicate that VLM-based evaluation offers strong semantic guidance for selecting behaviorally coherent trajectories. When both modules are enabled, MindDrive achieves the highest PDMS of 88.9 and the strongest overall performance. These results show that FaTG and VLoE offer complementary benefits: FaTG enhances the structural quality and physical plausibility of generated trajectories, while VLoE performs fine-grained semantic assessment, allowing MindDrive to reliably select safe and context-aware trajectory.

\subsubsection{Ablation Study on World Action Model}

\begin{table}
    \centering
    
\centering
\addtolength{\tabcolsep}{0.2pt}

\renewcommand{\arraystretch}{0.9} %
\resizebox{\linewidth}{!}{
\begin{tabular}{l|>{\columncolor{gray!30}}c|ccccccccc}
\toprule
Architecture & 
PDMS$\uparrow$ &
NC$\uparrow$ & 
DAC$\uparrow$ & 
EP$\uparrow$ & 
TTC$\uparrow$ & 
Comf.$\uparrow$   \\
\midrule
Mamba-Only                              & 85.6 & 97.6 & 94.6 & 80.4 & 93.1 & 99.7 \\
Hybrid-Structure\textsuperscript{1}     & 88.1 & 98.5 & 96.3 & 81.9 & 95.0 & \textbf{99.9} \\
Hybrid-Structure\textsuperscript{2}     & 88.5 & 98.4 & 96.6 & 82.5 & \textbf{95.1} & \textbf{99.9} \\
\midrule
Hybrid-Structure\textsuperscript{3}     & \textbf{88.9} & \textbf{98.4} & \textbf{97.0} & \textbf{82.8} & \textbf{95.1} & \textbf{99.9} \\

\bottomrule
\end{tabular}}
\caption{\textbf{Ablation Study on World Action Model architecture.} (On the Navtest split under the NAVSIM-v1 metrics.) 
Hybrid-Structure\textsuperscript{1/2/3} denotes Transformer–Mamba hybrid architectures:
1: a shallow Transformer–Mamba stack,
2: a symmetric Transformer–Mamba–Transformer design, and
3: a sandwich-style Transformer–2$\times$Mamba–Transformer design.}
\label{tab:ablation_wam_arch_PDMS}

    \vspace{-10pt}
\end{table}

We further investigate the design choices of the \textbf{world model} through two ablation studies: (1) different architectural configurations, and (2) different temporal prediction steps.

\noindent\textbf{Architectural Variants.} As shown in Table~\ref{tab:ablation_wam_arch_PDMS}, we evaluate four architectural variants of the world model. The pure Mamba model underperforms due to its limited ability to capture global spatial dependencies, leading to weaker predictive consistency. In contrast, the hybrid configurations yield substantial improvements, with the best performance achieved by the \textbf{Transformer–2$\times$Mamba–Transformer} variant, which effectively balances global attention with efficient temporal reasoning.

\begin{table}
    \centering
    
\centering
\renewcommand\arraystretch{0.9}
\resizebox{\linewidth}{!}{
\begin{tabular}{p{25mm}|>{\columncolor{gray!30}}c|ccccc}
\toprule
 \makecell[l]{Rollout Steps} & PDMS $\uparrow$ & NC $\uparrow$ & DAC $\uparrow$ & EP $\uparrow$ & TTC $\uparrow$ & Comf. $\uparrow$ \\ 
                \midrule
0s $\rightarrow$ 4s & 87.5 & 98.1 & 96.1 & 81.7 & 94.3 & \textbf{99.9}  \\ 
\midrule
0s $\rightarrow$ 2s $\rightarrow$ 4s & \textbf{88.9} & \textbf{98.4} & \textbf{97.0} & \textbf{82.8} & \textbf{95.1} & \textbf{99.9} \\ 

\bottomrule
\end{tabular}}
\caption{\textbf{Ablation Study on World Action Model time steps.} (On the Navtest split under the
NAVSIM-v1 metrics.)}
\label{tab:ablation_wam_steps}

    \vspace{-10pt}
\end{table}

\begin{table}
    \centering

\centering
\addtolength{\tabcolsep}{0.2pt}

\resizebox{\linewidth}{!}{
\begin{tabular}{l|>{\columncolor{gray!30}}c|ccccccccc}
\toprule
VLoE & 
PDMS$\uparrow$ &
NC$\uparrow$ & 
DAC$\uparrow$ & 
EP$\uparrow$ & 
TTC$\uparrow$ & 
Comf.$\uparrow$   \\
\midrule
Without VLM    & 64.5 & 97.6 & 61.0 & 59.4 & 93.1 & 99.7 \\
\midrule
With VLM       & \textbf{67.5} & \textbf{98.5} & \textbf{75.6} & \textbf{63.1} & \textbf{95.5} & \textbf{100}  \\

\bottomrule
\end{tabular}}
\caption{\textbf{Ablation Study on VLoE.} (On the Navsafe split under the NAVSIM-v1 metrics.) }
\label{tab:ablation_vloe_arch_PDMS}

    \vspace{-10pt}
\end{table}

\begin{table}
    \centering

\centering
\addtolength{\tabcolsep}{0.2pt}
\tabcolsep=0.5mm
\renewcommand{\arraystretch}{1.2} %
\resizebox{\linewidth}{!}{
\begin{tabular}{p{20mm}|>{\columncolor{gray!30}}c|ccccccccc}
\toprule
VLoE & 
EPDMS$_1$$\uparrow$ &
NC$\uparrow$ & 
DAC$\uparrow$ & 
DDC$\uparrow$ & 
TLC$\uparrow$ & 
EP$\uparrow$ & 
TTC$\uparrow$ & 
LK$\uparrow$ & 
HC$\uparrow$ & 
EC$\uparrow$   \\
\midrule

Without VLM  & 68.2 
& 96.1 & 85.1 & 98.5 & 99.3 & 80.6 & 93.7 & 95.1 & 97.3 & 70.2 \\
\midrule
With VLM     & \textbf{72.7} 
& \textbf{97.0} & \textbf{89.7} & \textbf{98.6} & \textbf{99.3} & \textbf{80.7} & \textbf{95.8} & \textbf{96.0} & \textbf{97.5} & \textbf{71.0} \\ 

\bottomrule
\end{tabular}}
\caption{\textbf{Ablation Study on VLoE.} (On the Navhard split under the
NAVSIM-v2 metrics.)  EPDMS$_1$ represents the EPDMS stage 1 evaluated without synthetic scenarios.}
\label{tab:ablation_wam_arch_EPDMS}

    \vspace{-10pt}
\end{table}

\noindent\textbf{Temporal Step Analysis.} We also conduct an ablation on the number of \textbf{future prediction steps}, as summarized in Table~\ref{tab:ablation_wam_steps}. The BEV world model predicts future states recurrently, where $A \rightarrow B$ denotes that the predicted state at time $B$ is based on the state at time $A$.
Comparing direct long-horizon prediction ($0\text{s}\rightarrow4\text{s}$) with recurrent short-step prediction ($0\text{s}\rightarrow2\text{s}\rightarrow4\text{s}$), the latter achieves consistently higher scores across all metrics (NC, DAC, EP, TTC, PDMS). This suggests that step-wise temporal unrolling improves temporal consistency and reduces compounding prediction errors in long-horizon forecasting.

\subsubsection{Ablation Study on VLoE Architecture}

To assess the contribution of the \textbf{VLoE} module, we perform ablations under both NAVSIM-v1 (PDMS on Navsafe) and NAVSIM-v2 (EPDMS$_1$ on Navhard) metrics. This allows us to evaluate whether the benefits of vision–language reasoning remain consistent across different scoring schemes. Across both protocols, adding the VLM yields clear improvements. On the Navsafe split, PDMS increases from 64.5 to \textbf{67.5}, with notable gains in NC, DAC, and EP (Table~\ref{tab:ablation_vloe_arch_PDMS}).

Under the stricter NAVSIM-v2 metrics, EPDMS$_1$ improves from 68.2 to \textbf{72.7}, alongside broader increases in DAC, TTC and LK (Table~\ref{tab:ablation_wam_arch_EPDMS}). These results demonstrate that language-guided reasoning in VLoE consistently enhances both safety and compliance, indicating benefits that extend beyond purely visual cues and generalize across evaluation settings.

\subsection{Visualization}

\begin{figure}[t]
\centering
 \includegraphics[width=1.0\linewidth]{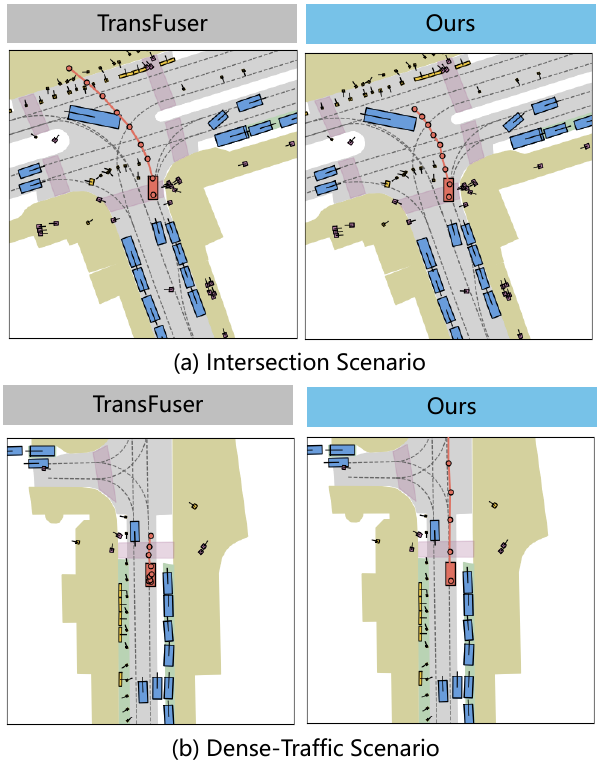}
\caption[ ]{Qualitative comparison between TransFuser~\cite{TransFuser} and our method on challenging Navsafe scenarios.
(a) Intersection: TransFuser exhibits trajectory deviations and unstable turning behavior, whereas our method produces smoother and lane-consistent plans.
(b) Dense-traffic: under heavy flow, TransFuser predictions become jittery and drift toward surrounding vehicles; while our method maintains stable, congestion-aware trajectories.}
\label{fig:vis}
\end{figure}

As shown in Figure~\ref{fig:vis}, we present a qualitative comparison between TransFuser~\cite{TransFuser} and our method on challenging scenarios from the Navsafe split. In intersection cases, TransFuser often produces drifting or unstable turning trajectories, whereas our model yields smoother and lane-consistent plans. In dense-traffic scenes, TransFuser frequently deviates toward surrounding agents, which can lead to direct collisions or high-risk near-misses. In contrast, our method maintains stable, congestion-aware trajectories that preserve safe distances and better reflect the underlying driving intent.

\section{Conclusion} \label{sec:conclusion}
Our MindDrive introduces a structured reasoning paradigm that integrates what-if simulation, trajectory candidate generation, and multi-objective evaluation into a harmonized E2E-AD framework. The proposed Future-aware Trajectory Generator (FaTG), driven by a World Action Model (WAM), enables ego-conditioned “what-if” scene rollouts for foresighted trajectory planning. Building on these predictive candidates, the VLM-oriented Evaluator (VLoE) leverages a vision–language model to assess safety, comfort, and efficiency, yielding decisions that align with human driving intent. Extensive experiments on NAVSIM-v1 and NAVSIM-v2 demonstrate that MindDrive achieves consistent improvements in safety, smoothness, and generalization compared to state-of-the-art E2E-AD approaches. Future work includes incorporating reinforcement learning into the world–VLM reasoning loop to enable continuous self-improvement and further enhance system robustness in complex environments.

\section*{ACKNOWLEDGMENTS} 
This work was supported in part by the National Key R\&D Program of China (Grant No. 2023YFB3107400) and the National Natural Science Foundation of China (Grant No. U22A2042), which contributed to both the study design and financial support of this research.

 \bibliographystyle{elsarticle-num} 
 \bibliography{ref}



\end{document}